\documentclass[conference]{IEEEtran}
\IEEEoverridecommandlockouts
\usepackage{cite}
\usepackage{amsmath,amssymb,amsfonts}
\usepackage{algorithmic}
\usepackage{graphicx}
\usepackage{textcomp}
\usepackage{xcolor}
\usepackage{multirow}
\usepackage{booktabs}
\def\BibTeX{{\rm B\kern-.05em{\sc i\kern-.025em b}\kern-.08em
    T\kern-.1667em\lower.7ex\hbox{E}\kern-.125emX}}
\begin{document}

\title{ISCS: Parameter-Guided Feature Pruning for Resource-Constrained Embodied Perception}

\author{Jinhao Wang\\
Santa Clara University\\
Santa Clara, CA, USA\\
{\tt\small jwang11@scu.edu}
\and
Nam Ling\\
Santa Clara University\\
Santa Clara, CA, USA\\
{\tt\small nling@scu.edu}
\and
Wei Wang\\
Futurewei Technologies, Inc.\\
San Jose, CA, USA\\
{\tt\small rickweiwang@futurewei.com}
\and
Wei Jiang\\
Futurewei Technologies, Inc.\\
San Jose, CA, USA\\
{\tt\small wjiang@futurewei.com}
}
\maketitle

\begin{abstract}
Prior studies in embodied AI consistently show that robust perception is critical for human-robot interaction, yet deploying high-fidelity visual models on resource-constrained agents remains challenging due to limited on-device computation power and transmission latency. Exploiting the redundancy in latent representations could improve system efficiency, yet existing approaches often rely on costly dataset-specific ablation tests or heavy entropy models unsuitable for real-time edge-robot collaboration. 

We propose a generalizable, dataset-agnostic method to identify and selectively transmit structure-critical channels in pretrained encoders. Instead of brute-force empirical evaluations, our approach leverages intrinsic parameter statistics—weight variances and biases—to estimate channel importance. This analysis reveals a consistent organizational structure, termed the Invariant Salient Channel Space (ISCS), where Salient-Core channels capture dominant structures while Salient-Auxiliary channels encode fine visual details. Building on ISCS, we introduce a deterministic static pruning strategy that enables lightweight split-computing. 

Experiments across different datasets demonstrate that our method achieves a deterministic, ultra-low latency pipeline by bypassing heavy entropy modeling. Our method reduces end-to-end latency, providing a critical speed-accuracy trade-off for resource-constrained human-aware embodied systems.
\end{abstract}

\begin{IEEEkeywords}
Embodied AI, Split Computing, Feature Pruning, Resource-constrained Perception
\end{IEEEkeywords}

\section{Introduction}
The evolution of embodied AI towards long-term human-agent cohabitation requires robust multimodal perception to ensure safety, trust, and effective collaboration. A consistent observation in this domain is that deploying high-fidelity visual models on mobile agents, such as service robots and autonomous drones, is often constrained by limited on-device computation power. While split-computing offers a viable paradigm by offloading heavy downstream tasks to edge servers, it introduces a critical bottleneck: the instability and transmission latency. Consequently, optimizing the trade-off between end-to-end responsiveness and perception accuracy is crucial\cite{Yao2025Energy-Efficient} for real-time, human-aware embodied systems.

While minimizing bandwidth is important, existing Learned Image Compression (LIC) methods over-optimize for rate-distortion at the expense of latency. They rely on serial autoregressive entropy models, which become a computational bottleneck on edge devices. For a robot operating in a dynamic environment, a delayed perception is as dangerous as a blind one. Our framework challenges this convention by prioritizing execution latency. We demonstrate that a modest increase in transmission payload (entropy-free packing) is a worthy trade-off for substantially faster, jitter-free inference.

In this paper, we propose a generalizable, dataset-agnostic framework for resource-constrained embodied AI. Instead of relying on costly data-dependent ablations, we exploit the model's own parameter statistics. Specifically, we hypothesize from a parameter distribution perspective that weight variances and bias magnitudes carry strong signals about channel importance. Channels with high weight variances tend to capture salient structural features, while biases often encode global priors. This analysis reveals a consistent organizational structure, we term this structure the Invariant Salient Channel Space (ISCS). Within each ISCS, a Salient-Core (SC) channel captures critical structural features, while the associated Salient-Auxiliary (SA) channels contribute complementary cues, such as fine color and texture details. For semantic tasks, structural information is paramount while texture detail is redundant. Building on this hypothesis, we introduce a static pruning strategy that physically retains SC channels. By purely relying on channel reduction and bypassing the computationally expensive entropy model, we achieve an entropy-free split-computing pipeline that ensures minimal latency.


\begin{figure*}[htp]
    \centering
    \includegraphics[width=1\linewidth]{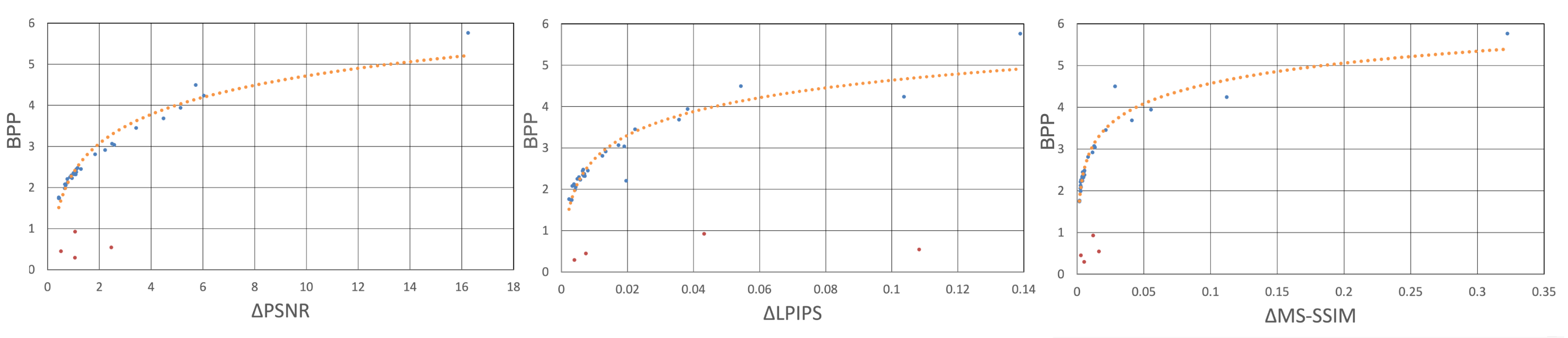}
    \vspace{-2em}
    \caption{Reconstruction loss from single-channel removal. The y-axis shows per-channel BPP estimated by the model’s probability model. Most channels (blue dots) exhibit a strong logarithmic correlation between per-channel BPP and reconstruction quality degradation. However, a small set of outlier channels (red dots) exist with only small BPP yet causing disproportionately large loss when removed. } 
    \vspace{-2mm}
    \label{fig:ch_ablation}
\end{figure*}

Our main contributions are as follows:
\begin{itemize}
    \item \textbf{Parameter-importance-based ISCS Discovery: }We identify structure-critical channels solely by examining the weight parameters of pretrained models. These statistics are intrinsic to the trained model and remain invariant across datasets, offering robustness and computational efficiency without the need for costly empirical ablations such as brute-force per-channel removal tests.
    \item \textbf{Entropy-free Deterministic Pruning: }Based on ISCS, we introduce a static pruning strategy that physically discards machine-redundant auxiliary channels. This design enables ultra-low latency transmission for edge-robot collaboration without addition entropy coding.
    \item \textbf{Modular Integration with Split-computing Framework: }Our method is modular and can be easily integrated with any pretrained split-computing frameworks. We demonstrate effectiveness across multiple downstream tasks, showing consistent latency reduction and computational speedup while preserving downstream accuracy. 
    \item \textbf{Human-aware Performance: }Experiments on human-centric datasets demonstrate that our method preserves high-fidelity features for human detection and action recognition.
\end{itemize}

\section{Related Work}
\subsection{Feature Redundancy in Deep Networks}
Deep neural networks typically operate in high-dimensional latent spaces, yet empirical studies suggest that effective information lies on a much lower-dimensional manifold\cite{Bhoopalan2025Task}. A consistent observation across domains is that representations are often sparse or highly correlated. In the context of Learned Image Compression (LIC), this implies uneven information distribution: a small subset of channels captures the majority of the rate-distortion energy. While recent works attempt to identify these channels via data-dependent ablations, they often retain the entropy model to minimize bitrate. As noted in hardware studies, the serial nature of entropy decoding remains a primary latency bottleneck, limiting practical acceleration on edge devices\cite{Bai2025EdgeMM:}.

\subsection{Compression for Machine Perception}
Standard codecs, such as JPEG, optimized for human vision, introduce blocking artifacts at low bitrates that degrade the structural integrity required by machine vision. To address this, Video Coding for Machines (VCM) has emerged. Recent scalable frameworks, such as the task-driven coding by Choi \& Bajić \cite{Choi_2022} and DeepFGS \cite{DeepFGS}, optimize features to support variable bitrates. Others, like Adapt-ICMH \cite{AdaptICMH}, employ adapter-based mechanisms to tune features for specific tasks. While superior in rate-accuracy trade-offs compared to standard codecs, these methods typically retain complex entropy modeling, or require transmitting additional side information. This introduces non-deterministic decoding latency, which is often prohibitive for hard real-time robotic scenario.

\subsection{Split Computing and Low-Latency Inference}
Split computing offloads computational burdens from resource-constrained agents to edge servers\cite{Naveen2021LOW}. Existing frameworks like TOFC \cite{TOFC} and dynamic strategies like AVERY \cite{AVERY} adaptively partition computation based on network conditions\cite{Jain2021Latency-Memory}. However, these mechanisms often incur runtime decision overhead. Furthermore, even "low-latency" entropy models proposed in recent studies (e.g., DCAE \cite{DCAE}) still necessitate serial bitstream parsing.In contrast, our ISCS framework prioritizes execution latency. We propose a deterministic, \textit{entropy-free} pipeline driven by intrinsic parameter statistics. By replacing serial decoding with raw INT8 packing and static pruning, we transform the edge-server reconstruction into a fully parallel operation with constant step complexity $\mathcal{O}(1)$, decoupling latency from image complexity.

\section{Methodology}
\begin{figure*}[htp]
    \centering
    \includegraphics[width=0.95\linewidth]{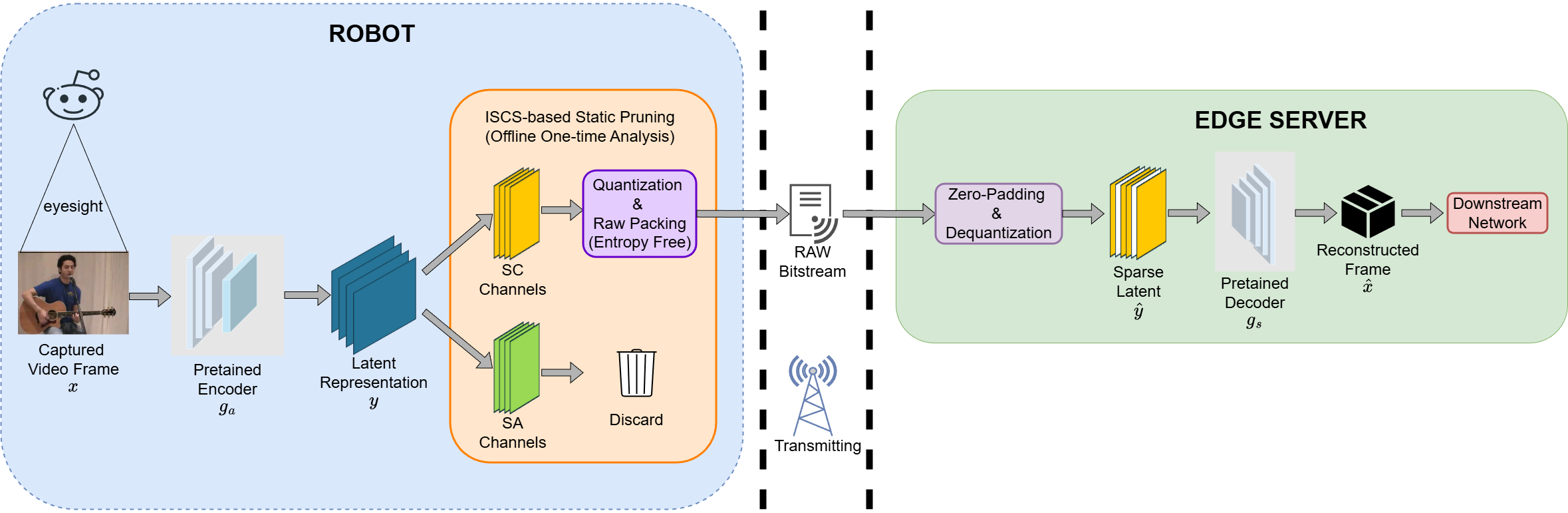}\vspace{-.9em}
    \caption{Overview of the proposed Entropy-free Split-Computing framework. The resource-constrained Robot (Client) encodes the visual input and performs ISCS-based Static Pruning, physically discarding machine-redundant SA channels while retaining structure-critical SC channels. To minimize latency, the selected features are transmitted via a lightweight raw packing protocol, bypassing the computationally expensive entropy model. The Edge Server receives the raw bitstream and reconstructs the sparse latent representation $\hat{y}$ via zero-padding for downstream perception tasks.}
    \label{fig:mainflow}\vspace{-1em}
\end{figure*}
We integrate our method into a pretrained MLIC\cite{MLIC++} framework to enable efficient split-computing. As shown in Figure \ref{fig:mainflow}, we keep the encoder $g_a$ intact but analyze its final projection weights to identify the ISCS structure. Based on this one-time analysis, we perform a deterministic static pruning to physically select SC channels before transmission, thereby bypassing the entropy bottleneck.

\subsection{Per-Channel Importance Ablations}\label{sec:ablation}

The most intuitive way to measure channel importance is from an information-importance perspective. When channels are treated independently, those with higher bitrates exhibit larger entropy and are expected to contribute more to reconstruction quality. Consequently, removing such channels should cause greater reconstruction loss. To quantify the importance of a channel $c$, we perform a straightforward channel-removal test and measure the degradation using three metrics: $\Delta$PSNR, $\Delta$MS-SSIM, and $\Delta$LPIPS (the smaller the better):
\begin{eqnarray*}
    \Delta \text{PSNR}_c &=& \text{PSNR}_{\text{baseline}} - \text{PSNR}_{c=0},\nonumber\\
    \Delta \text{LPIPS}_c &=& \text{LPIPS}_{c=0} - \text{LPIPS}_{\text{baseline}},\nonumber\\
    \Delta \text{MS\text{-}SSIM}_c &=& \text{MS\text{-}SSIM}_{\text{baseline}} - \text{MS\text{-}SSIM}_{c=0}
\end{eqnarray*}
We compute the performance drop between baseline reconstruction and the reconstruction with channel $c$ removed ($c\!=\!0$). Figure \ref{fig:ch_ablation} reports results on JPEG\_AI test set. Across all metrics, we observe a consistent positive correlation between per-channel bitrate and reconstruction loss.

However, the results also reveal several outlier channels—those with small bitrate but substantial impact on reconstruction quality when removed, highlighting the limitations of relying on per-channel bitrate alone to predict channel importance.

\subsection{Parameter Importance and ISCS}\label{sec:ISCS}

\begin{figure*}[th]
    \centering
    \includegraphics[width=\linewidth]{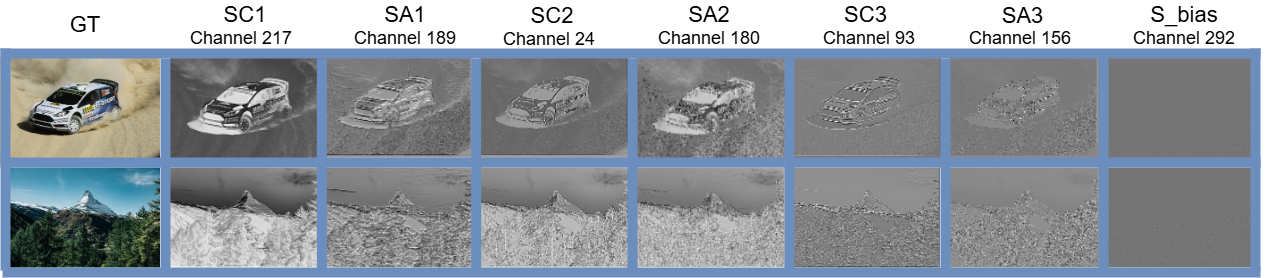}
    \caption{Visual examples of per-channel latent activations for different types of channels in ISCS. The SC channels clearly capture prominent structural patterns. Their corresponding SA channels provide complementary cues, such as fine visual details. In contrast, bias-dominated channels exhibit near-constant activations across different image inputs.}\vspace{-1em}
    \label{fig:semantic}
\end{figure*}

We define ISCS as the latent structure discovered by analyzing the encoder’s final projection weights, where each Salient-Core (SC) channel acts as an anchor surrounded by Salient-Auxiliary (SA) channels that carry correlated or complementary information, \textit{e.g.}, textures, colors, fine details (some visualization examples shown in Figure \ref{fig:semantic}). 

Assume the encoder’s final linear transformation projects inputs into latent channels ${z_c}$. Each output channel $c$ is parameterized by a kernel tensor $W_c \in \mathbb{R}^{C_{\mathrm{in}}\times K \times K}$ and, optionally, a scalar bias $b_c$. From ${W_c, b_c}$ we compute three types of importance scores to identify ISCS.

\subsubsection{Variance score for identifying SCs}\label{para:S1}

High weight variance implies a broader linear response and a larger activation dynamic range under typical inputs, which aligns with higher entropy and stronger rate-distortion contribution. Channels with high variance scores therefore serve as SCs, capturing critical structural information. The variance score of channel $c$ is defined as:
\begin{equation}
\operatorname{Var}(W_c) = \frac{1}{K^2 C_{\mathrm{in}}} \sum_{i,j,k} \bigl( W_c[i,j,k] - \mu_c \bigr)^2,\label{eq:variance}
\end{equation}

\noindent with mean 

\begin{equation}
    \mu_c = \frac{1}{K^2 C_{\mathrm{in}}} \sum_{i,j,k} W_c[i,j,k].\nonumber
\end{equation}
\noindent Here, $W_c[i,j,k]$ indexes input channel $i$ at spatial location $(j,k)$, $C_{\mathrm{in}}$ is the number of input channels to the layer, $K$ is the kernel size, and $\mu_c$ and $\operatorname{Var}(W_c)$ are the entry-wise mean and variance of $W_c$ respectively.
Channels with high $\operatorname{Var}(W_c)$ produce activations with larger dynamic range, often capturing salient structural information, and therefore are treated as SCs.

\subsubsection{Cosine similarity for identifying SA channels}
SA channels are defined as those strongly correlated with a given SC. We measure inter-channel correlation using the pairwise cosine similarity between flattened kernel vectors:

\begin{equation}
\mathrm{Sim}(\bar{W}c, \bar{W}{c'}) = \frac{\langle \bar{W}c, \bar{W}{c'} \rangle}{|\bar{W}c|,|\bar{W}{c'}|},
\end{equation}

\noindent where $\bar{W}_c$ is the flattened vector of kernel tensor $W_c$. 

SA channels typically encode correlated fine details or textural refinements. While beneficial for pixel-perfect human visualization, we hypothesize that these SA channels are semantically redundant for machine vision tasks, which rely primarily on the structural backbone provided by SCs.

\subsubsection{Bias score for additional important channels}\label{para:S3}
Large biases often encode input-invariant global priors, such as luminance or overall tone. Channels with large bias scores $S_{\mathrm{bias}}$ are therefore identified as highly important:

\begin{equation}
    S_{\mathrm{bias}}(c) = |b_c|.\label{eq:bias}
\end{equation}

\noindent Interestingly, these bias-dominated channels correspond to the outliers observed in earlier empirical tests—those with very small per-channel bitrate but disproportionately large reconstruction loss when removed.

In practice, the ISCS is identified through a one-time weight parameter analysis: channels are first ranked by variance score $\operatorname{Var}(W_c)$, and top channels are used as SCs $S_{\mathrm{centrality}}$; for each SC channel $c$, non-SC channels are ranked based on their cosine similarity $\mathrm{Sim}(\bar{W}c,\bar{W}{c'})$; finally, additional bias-dominated channels are incorporated using $S_{\mathrm{bias}}$. Because these scores depend only on model parameters, they are input- and dataset-invariant, making ISCS a stable foundation for our proposed split-computing pipeline.

\subsubsection{Visualization of ISCS channels}
Figure \ref{fig:semantic} shows visual examples of per-channel latent activations for different types of channels in the discovered ISCS of a pretrained MLIC+\cite{MLIC++} model. The top-ranked SC channels clearly capture prominent structural patterns that are most critical for reconstruction. Their corresponding SA channels (the most correlated ones shown) provide complementary cues, such as fine visual details. In contrast, bias-dominated channels exhibit near-constant activations across different image inputs, suggesting that their bitrate can be further reduced by transmitting a per-channel scalar instead of a full latent feature for such channels.

Our extensive empirical validation across multiple datasets (Sec.~\ref{sec:exper}) confirms it as a robust heuristic for resource-constrained engineering.

\subsection{Entropy-free Static Pruning}
Based on the discovered ISCS structure, we propose a static pruning strategy tailored for real-time split-computing. Unlike traditional approaches that retain all channels and rely on entropy coding to compress and transmit them, our method physically reduces the feature dimensionality to achieve low-latency transmission.

\subsubsection{Pruning Strategy}
Let $\mathcal{C}_{total}$ be the set of all latent channels. We compute a composite importance rank for each channel based on the statistics defined in Sec.~\ref{sec:ISCS}. Specifically, we prioritize Salient-Core (SC) channels (high variance) for structural definition and Bias-Dominated channels (high bias) for feature distribution stability.
We select the top-$k$ channels to form the transmission set $\mathcal{C}_{trans}$, while the remaining set $\mathcal{C}_{discard}$, which consists primarily of machine-vision redundant SA channels, is permanently pruned:
$$\hat{y}_c = \begin{cases} y_c, & \text{if } c \in \mathcal{C}_{trans} \\ 0, & \text{if } c \in \mathcal{C}_{discard} \end{cases}$$
By discarding $\mathcal{C}_{discard}$ channels, we effectively filter out high-frequency textural noise that is less relevant for semantic inference, significantly reducing the data volume.


\subsubsection{Embodied Split-Computing Protocol}
In order to achieve long-term human–agent cohabitation, we implement the proposed method as a lightweight split-computing protocol. This protocol shifts the computational burden from the resource-constrained robot to the edge server. The workflow proceeds as follows:
\begin{itemize}
    \item \textbf{On-Device:} The mobile agent captures the video frame and passes it through the pretrained encoder $g_a$. Instead of running a computationally heavy entropy model, the agent simply extracts the indices corresponding to $\mathcal{C}_{trans}$. These selected feature maps are directly quantized to INT8 and packed, bypassing the entropy model entirely. This reduces the on-device computational complexity to a simple memory copy operation.
    \item \textbf{Transmission:}  The removal of the entropy bottleneck ensures that transmission latency is determined solely by bandwidth and packet size, avoiding serialization delays.
    \item \textbf{On-Server:} The edge server receives the bitstream and reconstructs the tensor by scattering the received data into their original indices and zero-padding the pruned indices $\mathcal{C}_{discard}$. The reconstructed feature map $\hat{y}$ is then fed into the decoder $g_s$ and subsequent downstream task models. 
\end{itemize}
This asymmetrical design ensures that the robot maintains minimal inference latency and power consumption, enabling sustained real-time operation in dynamic environments.

\begin{table*}[htbp]
\centering
\renewcommand{\arraystretch}{0.85}
\caption{\textbf{Quantitative comparison on downstream tasks.} We report mAP for Object Detection (COCO) and Top-1 Accuracy for Action Recognition (UCF101). At 25\% retention, our method achieves performance comparable to the uncompressed Original, verifying the effectiveness of ISCS-based pruning.}
\vspace{-0.2cm}
\begin{tabular}{ccccccccc}
\toprule
\textbf{Metric} & \multicolumn{8}{c}{\textbf{Method}} \\
\cmidrule(lr){2-9}
\textbf{Object Detection} & \textbf{\textit{Original}} & \textbf{\textit{5\%}} & \textbf{\textit{20\%}} & \textbf{\textit{25\%}} & \textbf{\textit{40\%}} & \textbf{\textit{Random(25\%)}} & \textbf{\textit{Reverse(25\%)}} & \textbf{\textit{JPEG}} \\
\midrule
mAP50    & 0.745 & 0.633 & 0.736 & 0.741 & 0.751 & 0.495 & 0.363 & 0.744  \\
mAP50-95 & 0.514 & 0.415 & 0.507 & 0.513 & 0.523 & 0.297 & 0.204 & 0.514  \\
\midrule
\textbf{Action Recognition} & \multicolumn{8}{c}{} \\
\midrule
Best Prec@1 & 89.85 & 78.201 & 88.12 & 88.581 & 87.313 & 60.323 & 40.715 & 89.313 \\
\bottomrule
\end{tabular}
\label{tab1}
\vspace{-0.4cm}
\end{table*}

\begin{table*}[htbp]
\centering
\renewcommand{\arraystretch}{0.85}
\caption{Time comparison on Kodak and JPEG-AI. Both overall runtime and the isolated runtime of context model are reported.}
\vspace{-0.2cm}
\begin{tabular}{cccccccccc}
\toprule
\multirow{3}{*}{\textbf{Model}} & \multicolumn{4}{c}{\textbf{Overall Runtime}} & \multicolumn{4}{c}{\textbf{Context Modules Only}} & \multirow{3}{*}{\textbf{BPP}} \\ 
\cmidrule(lr){2-5} \cmidrule(lr){6-9}
 & \multicolumn{2}{c}{Encoding Time (s)} & \multicolumn{2}{c}{Decoding Time (ms)} & \multicolumn{2}{c}{Encoding Time (s)} & \multicolumn{2}{c}{Decoding Time (ms)} & \\
\cmidrule(lr){2-3} \cmidrule(lr){4-5} \cmidrule(lr){6-7} \cmidrule(lr){8-9}
 & Kodak & JPEG-AI & Kodak & JPEG-AI & Kodak & JPEG-AI & Kodak & JPEG-AI & \\
\midrule
MLIC+ & 9.912 & 13.412 & 74.2 & 265.9 & 5.322 & 6.201 & 35.8 & 115.7 & 0.3 \\
Ours  & 0.453 & 7.275  & 3.397& 6.917 & 0     & 0     & 0    & 0     & 0.5 \\
\bottomrule
\end{tabular}
\label{tab:efficiency}
\end{table*}

\section{Experiments}\label{sec:exper}
We evaluate the proposed ISCS-based pruning framework in the context of resource-constrained embodied perception. We aim to demonstrate two key claims:
\begin{itemize}
    \item \textbf{High-Fidelity Perception: }Our method retains critical structural semantics for machine vision tasks even at extreme compression rates.
    \item \textbf{Efficiency: }The entropy-free design eliminates the decoding latency bottleneck, making it suitable for real-time edge-robotic collaboration.
\end{itemize}

\subsection{Experimental Setup}\label{sec:setting}
\subsubsection{Datasets and Downstream Tasks} 
To simulate human-aware embodied scenarios, we conduct experiments on two representative datasets:
\begin{itemize}
    \item \textbf{COCO 2017 (Human Detection):} We evaluate object detection performance on the "Person" class of the COCO validation set, using a pretrained YOLOv8n as the downstream task model. This simulates a robot detecting humans in static scenes.
    \item \textbf{UCF101 (Action Recognition):} We use the UCF101 dataset to evaluate human action recognition. A Temporal Shift Module (TSM) model is employed as the downstream backbone. This simulates an agent understanding human behavior in dynamic video streams.
\end{itemize}

\subsubsection{Baseline \& Ablation}
\begin{itemize}
    \item \textbf{Original:} The performance of downstream models on uncompressed raw images.
    \item \textbf{Random Pruning:} Channels are randomly selected to match the target data volume, serving as a baseline to validate the effectiveness of our ISCS selection strategy.
    \item \textbf{Reverse Pruning:} Specifically selects the least important channels based on the ISCS ranking. This serves as a negative control to validate that SC channels indeed carry the essential semantic information.
    \item \textbf{JPEG:} We test JPEG as the standard codec baseline.
\end{itemize}

\subsubsection{Implementation Details}
We use a pretrained MLIC+ model as the feature extractor. All experiments are conducted on a single NVIDIA L40S GPU. We evaluate channel selection rates $\rho \in \{5\%, 20\%, 25\%, 40\%\}$ to cover various on-device constraints. For the Random Pruning baseline, we randomly select the same number of channels as our method to ensure a fair comparison under identical bandwidth budgets. The reverse pruning selects the least important channels (mainly SA channels) based on the ISCS ranking. The JPEG quality is set to 95, which has a similar bandwidth budget as the 25\% setting. To simulate our entropy-free transmission, the selected latent features ($\mathcal{C}_{trans}$) are quantized to 8-bit integers (INT8) to minimize bandwidth usage without additional entropy coding.

\begin{figure*}[htp]
    \centering
    \includegraphics[width=0.95\linewidth]{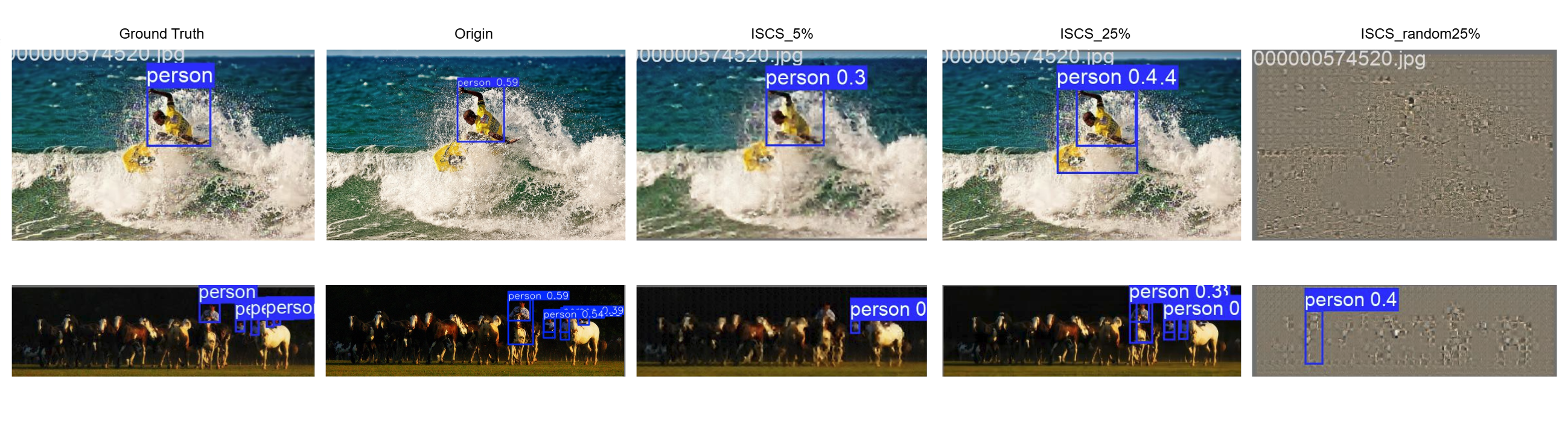}
    \vspace{-0.4cm}
    \caption{\textbf{Qualitative robustness comparison.} Our ISCS method preserves critical structural semantics, enabling accurate human detection even at high pruning. In contrast, the Random baseline suffers from catastrophic feature collapse.}
    \label{fig:quant}\vspace{-1em}
\end{figure*}

\subsection{Quantitative Results}


\subsubsection{Human Detection on COCO}
We first evaluate the impact of our method on static human detection. As presented in Table~\ref{tab1}, our method demonstrates exceptional robustness compared to both random pruning strategies and standard image codecs.

As shown in Table~\ref{tab1}, our method demonstrates exceptional robustness compared to non-pruning or random pruning strategies.
\begin{itemize}
    \item \textbf{High Fidelity at Low Bitrate:} At a 25\% channel retention rate, our ISCS-based method achieves an mAP@50 of 0.741, which is statistically indistinguishable from the uncompressed Original baseline ($0.745$) and the standard JPEG codec ($0.744$). This indicates that the removed 75\% of data was indeed redundant for the detection task.
    \item \textbf{Superiority over Baselines:} In contrast, the \textbf{Random} selection at the same 25\% budget suffers a significant performance drop to $0.495$ mAP. Crucially, the \textbf{Reverse} baseline yields the lowest score of $0.363$. This sharp degradation confirms our hypothesis that structural information is concentrated in the SC channels identified by ISCS.
    \item \textbf{Extreme Compression:} Even under an extreme constraint of 5\% retention, our method maintains a usable mAP of $0.633$.
\end{itemize}

\subsubsection{Action Recognition on UCF101}
Next, we assess temporal understanding on UCF101. Since action recognition relies on motion cues and temporal consistency, this task challenges the structural stability of the transmitted features.

The results in Table~\ref{tab1}on UCF101 follow a similar trend, further validating the generalizability of our framework.
\begin{itemize}
    \item \textbf{Preserving Temporal Semantics:} With 25\% of channels retained, our method maintains a Top-1 Accuracy of 88.58\%, closely matching the Original baseline of $89.85\%$, and remaining competitive with JPEG ($89.31\%$). This suggests that the SC channels capture not only spatial structures but also the essential features required for temporal reasoning, such as motion boundaries.
    \item \textbf{Impact of Structural Loss:} The \textbf{Random} ($60.32\%$) and \textbf{Reverse} ($40.71\%$) baselines exhibit catastrophic failure. This highlights that SA channels are insufficient for dynamic action recognition, and maintaining the spatial coherence of SC channels is vital for effective human-robot interaction tasks.
\end{itemize}

\subsubsection{Efficiency and Latency Analysis}

Unlike standard LIC methods that optimize solely for Rate-Distortion, our framework prioritizes the strict latency constraints of the embodied AI. We conducted a comprehensive benchmark simulating a real-world split computing setup: encoding is performed on an AMD EPYC 7513 CPU, simulating the resource-constrained client, while decoding is executed on a single Tesla H100 GPU, simulating the edge server.

\begin{itemize}
    \item \textbf{Client-Side Encoding:} As shown in Table~\ref{tab:efficiency}, the standard MLIC+ encoder is prohibitively slow for CPU execution, requiring 9.91s per frame on Kodak. This latency stems from the heavy computational load of the analysis transform and entropy estimation. In contrast, our pipeline reduces the encoding time to 0.45s per frame. By bypassing the entropy bottleneck and performing static channel pruning, we achieve a significant speedup on the client side, making resource-constrained inference feasible for mobile robots.
    
    \item \textbf{Server-Side Decoding:} On the server side, the baseline MLIC+ decoder is restricted by the serial nature of the autoregressive context model. Our entropy-free design unlocks the full parallel potential of the hardware, slashing the decoding latency to just 3.4ms. This deterministic, millisecond-level response is critical for closing the control loop in high-frequency robotic tasks.
    
    \item \textbf{BPP vs. Latency Trade-off:} While our method increases the bit-rate (0.5 bpp is reported on the 5\% pruning method), this is a deliberate trade-off. The modest increase in transmission payload is outweighed by the reduction in computational latency at both ends of the pipeline, ensuring the system meets real-time requirements. 
\end{itemize}


\subsection{Qualitative Visualization}
Figure \ref{fig:quant} visualizes the detection results on reconstructed inputs. Even at 5\% retention, our method retains critical structural outlines, enabling accurate detection. In sharp contrast, the Random method results in feature collapse even at a 5$\times$ higher budget (25\%). Notably, the second row illustrates the behavior of ISCS in complex scenes. While the Original model detects both the foreground persons and background objects, our method focuses on the dominant structural features. This suggests that ISCS tends to preserve the most salient visual information while reducing background redundancy, effectively minimizing bandwidth for the primary subject.

\section{Conclusion}
In this paper, we proposed a dataset-agnostic framework for embodied perception, driven by the hypothesis that structural information is paramount while texture is redundant for machine vision. By analyzing weight variances, bias magnitudes, and inter-channel correlations, we revealed a consistent ISCS structure, where structure-critical SC channels can be identified without costly ablations. Leveraging ISCS, our entropy-free static pruning strategy successfully eliminates the entropy bottleneck inherent in standard learned image compression. 

Extensive experiments across different downstream tasks confirm that our ISCS-based pruning preserves the detection and recognition precision required for robust human-agent interaction. This work provides a feasible, low-latency paradigm for real-time split-computing. Future directions include developing adaptive selection mechanisms that dynamically adjust the transmission set based on real-time channel conditions or task uncertainty, further optimizing the rate-distortion-latency trade-off.

\bibliographystyle{IEEEbib}
\bibliography{icme2026references}

@article{MLIC++,title={MLIC++: Linear Complexity Multi-Reference Entropy Modeling for Learned Image Compression},author={Wei Jiang and Jiayu Yang and Yongqi Zhai and Feng Gao and Ronggang Wang},journal={ACM Transactions on Multimedia Computing, Communications and Applications},year={2023},volume={21},pages={1 - 25},doi={10.1145/3719011}}

@article{Choi_2022,
   title={Scalable Image Coding for Humans and Machines},
   volume={31},
   ISSN={1941-0042},
   url={http://dx.doi.org/10.1109/TIP.2022.3160602},
   DOI={10.1109/tip.2022.3160602},
   journal={IEEE Transactions on Image Processing},
   publisher={Institute of Electrical and Electronics Engineers (IEEE)},
   author={Choi, Hyomin and Bajic, Ivan V.},
   year={2022},
   pages={2739–2754} }

@misc{DeepFGS,
      title={DeepFGS: Fine-Grained Scalable Coding for Learned Image Compression}, 
      author={Yi Ma and Yongqi Zhai and Ronggang Wang},
      year={2022},
      eprint={2201.01173},
      archivePrefix={arXiv},
      primaryClass={eess.IV},
      url={https://arxiv.org/abs/2201.01173}, 
}

@misc{AdaptICMH,
      title={Image Compression for Machine and Human Vision with Spatial-Frequency Adaptation}, 
      author={Han Li and Shaohui Li and Shuangrui Ding and Wenrui Dai and Maida Cao and Chenglin Li and Junni Zou and Hongkai Xiong},
      year={2024},
      eprint={2407.09853},
      archivePrefix={arXiv},
      primaryClass={cs.CV},
      url={https://arxiv.org/abs/2407.09853}, 
}

@misc{AVERY,
      title={AVERY: Adaptive VLM Split Computing through Embodied Self-Awareness for Efficient Disaster Response Systems}, 
      author={Rajat Bhattacharjya and Sing-Yao Wu and Hyunwoo Oh and Chaewon Nam and Suyeon Koo and Mohsen Imani and Elaheh Bozorgzadeh and Nikil Dutt},
      year={2025},
      eprint={2511.18151},
      archivePrefix={arXiv},
      primaryClass={cs.DC},
      url={https://arxiv.org/abs/2511.18151}, 
}

@article{TOFC,
   title={Task-Oriented Feature Compression for Multimodal Understanding via Device-Edge Co-Inference},
   ISSN={2161-9875},
   url={http://dx.doi.org/10.1109/TMC.2025.3626724},
   DOI={10.1109/tmc.2025.3626724},
   journal={IEEE Transactions on Mobile Computing},
   publisher={Institute of Electrical and Electronics Engineers (IEEE)},
   author={Yuan, Cheng and Liu, Zhening and Lv, Jiashu and Shao, Jiawei and Jiang, Yufei and Zhang, Jun and Li, Xuelong},
   year={2025},
   pages={1–14} }

@misc{DCAE,
      title={DC-AE 1.5: Accelerating Diffusion Model Convergence with Structured Latent Space}, 
      author={Junyu Chen and Dongyun Zou and Wenkun He and Junsong Chen and Enze Xie and Song Han and Han Cai},
      year={2025},
      eprint={2508.00413},
      archivePrefix={arXiv},
      primaryClass={cs.CV},
      url={https://arxiv.org/abs/2508.00413}, 
}

@article{Yao2025Energy-Efficient,
title={Energy-Efficient Edge Inference in Integrated Sensing, Communication, and Computation Networks},
author={Jiacheng Yao and Wei Xu and Guangxu Zhu and Kaibin Huang and Shuguang Cui},
journal={IEEE Journal on Selected Areas in Communications},
year={2025},
volume={43},
pages={3580-3595},
doi={10.1109/jsac.2025.3574612}
}

@article{Naveen2021LOW,
title={LOW LATENCY DEEP LEARNING INFERENCE MODEL FOR DISTRIBUTED INTELLIGENT IOT EDGE CLUSTERS},
author={Soumyalatha Naveen and Manjunath R. Kounte and Mohammed Riyaz Ahmed},
journal={IEEE Access},
year={2021},
volume={PP},
pages={1-1},
doi={10.1109/access.2021.3131396}
}

@article{Bai2025EdgeMM:,
title={EdgeMM: Multi-Core CPU with Heterogeneous AI-Extension and Activation-aware Weight Pruning for Multimodal LLMs at Edge},
author={Kangbo Bai and Le Ye and Ru Huang and Tianyu Jia},
journal={2025 62nd ACM/IEEE Design Automation Conference (DAC)},
year={2025},
pages={1-7},
doi={10.1109/dac63849.2025.11132644}
}

@article{Jain2021Latency-Memory,
title={Latency-Memory Optimized Splitting of Convolution Neural Networks for Resource Constrained Edge Devices},
author={Tanmay Jain and Avaneesh and Rohit Verma and R. Shorey},
journal={2022 14th International Conference on COMmunication Systems \& NETworkS (COMSNETS)},
year={2021},
pages={531-539},
doi={10.1109/comsnets53615.2022.9668356}
}

@article{Bhoopalan2025Task,
title={Task optimized vision transformer for diabetic retinopathy detection and classification in resource constrained early diagnosis settings},
author={Ramasubramanian Bhoopalan and Priyadharshini Sekar and N. Nagaprasad and Tadesse Regassa Mamo and R. Krishnaraj},
journal={Scientific Reports},
year={2025},
volume={15},
doi={10.1038/s41598-025-25399-1}
}

\vspace{12pt}

\end{document}